\documentclass[letterpaper, 10 pt, conference]{ieeeconf}  

\IEEEoverridecommandlockouts                              
\usepackage{verbatim}
\usepackage{mathtools}
\usepackage{amssymb}
\usepackage{enumerate}
\usepackage{tikz}
\usepackage{tikzscale}
\usepackage[hidelinks]{hyperref}
\usepackage{footmisc}
\usepackage{graphicx}
\usepackage{array}
\usepackage{textcomp}
\usepackage{gensymb}
\usepackage{cite}
\usepackage{float}
\usepackage{adjustbox}
\usepackage{diagbox}
\usepackage{units}
\usepackage[caption=false,font=footnotesize]{subfig}
\usetikzlibrary{positioning}
\usetikzlibrary{shapes.geometric, arrows}
\let\savedegree\degree
\let\degree\relax
\usepackage{mathabx}
\usepackage{booktabs}
\usepackage{microtype}
\let\degree\savedegree

\newcommand{\sevensensesymbol}{2}
\newcommand{\aslsymbol}{1}

\tikzstyle{startstop} = [rectangle, rounded corners, minimum width=3cm, minimum height=1cm,text centered, draw=black, fill=red!30]
\tikzstyle{io} = [trapezium, trapezium left angle=70, trapezium right angle=110, minimum width=0cm, minimum height=0cm, text centered, draw=black, fill=blue!30]
\tikzstyle{process} = [rectangle, minimum width=0cm, minimum height=0cm, text centered, text width=3cm, draw=black, fill=orange!30]
\tikzstyle{decision} = [diamond, minimum width=0cm, minimum height=0cm, text centered, draw=black, fill=green!30]
\tikzstyle{arrow} = [thick,->,>=stealth]

\title{\LARGE \bf
Dynamic Object Aware LiDAR SLAM based on Automatic Generation of Training Data
}

\author{Patrick Pfreundschuh$^{\aslsymbol,\sevensensesymbol}$, Hubertus F.C. Hendrikx$^{\sevensensesymbol}$, Victor Reijgwart$^{\aslsymbol}$, Renaud Dub\'e$^{\sevensensesymbol}$\\ Roland Siegwart$^{\aslsymbol}$, Andrei Cramariuc$^{\aslsymbol}$ \\

\\ \small$^{\aslsymbol}$Autonomous Systems Lab, ETH Z\"{u}rich, {\tt\footnotesize \{firstname.lastname\}@mavt.ethz.ch}

\\ \small$^{\sevensensesymbol}$Sevensense Robotics AG, {\tt\footnotesize \{firstname.lastname\}@sevensense.ch}

}

\begin{document}

\maketitle
\thispagestyle{empty}
\pagestyle{empty}

\begin{abstract}
Highly dynamic environments, with moving objects such as cars or humans, can pose a performance challenge for LiDAR SLAM systems that assume largely static scenes.
To overcome this challenge and support the deployment of robots in real world scenarios, we propose a complete solution for a dynamic object aware LiDAR SLAM algorithm.
This is achieved by leveraging a real-time capable neural network that can detect dynamic objects, thus allowing our system to deal with them explicitly.
To efficiently generate the necessary training data which is key to our approach, we present a novel end-to-end occupancy grid based pipeline that can automatically label a wide variety of arbitrary dynamic objects.
Our solution can thus generalize to different environments without the need for expensive manual labeling and at the same time avoids assumptions about the presence of a predefined set of known objects in the scene. 
Using this technique, we automatically label over 12000 LiDAR scans collected in an urban environment with a large amount of pedestrians and use this data to train a neural network, achieving an average segmentation IoU of $\mathbf{0.82}$.
We show that explicitly dealing with dynamic objects can improve the LiDAR SLAM odometry performance by $\mathbf{39.6\%}$ while yielding maps which better represent the environments. A supplementary video\footnote{\url{https://youtu.be/LcDxd97r1Gc}} as well as our test data\footnote{\url{https://projects.asl.ethz.ch/datasets/doals}} are available online.
\end{abstract}

\section{Introduction}

Simultaneous Localization and Mapping (SLAM) is an important capability of autonomous robots~\cite{cesar}.
Feature-based LiDAR SLAM algorithms often use matches between currently and previously extracted 3D features to estimate the trajectory of a robot~\cite{loam,dube2017online,legoloam}.
Such approaches typically assume that the majority of features are fixed in space.
However, in many cases robots are exposed to dynamic environments where the presence of moving objects is unavoidable, e.g. urban delivery robots operating among pedestrians.
Geometric features extracted from such dynamic objects add uncertainty and thus can lead to increased inaccuracy in the pose estimation~\cite{highway}. 
Dynamic objects can also end up in 3D reconstructions created by SLAM algorithms, which is not desired since these features are likely to no longer exist when the same places are revisited.
\begin{figure}[t]
\centering
\includegraphics[width=0.49\linewidth]{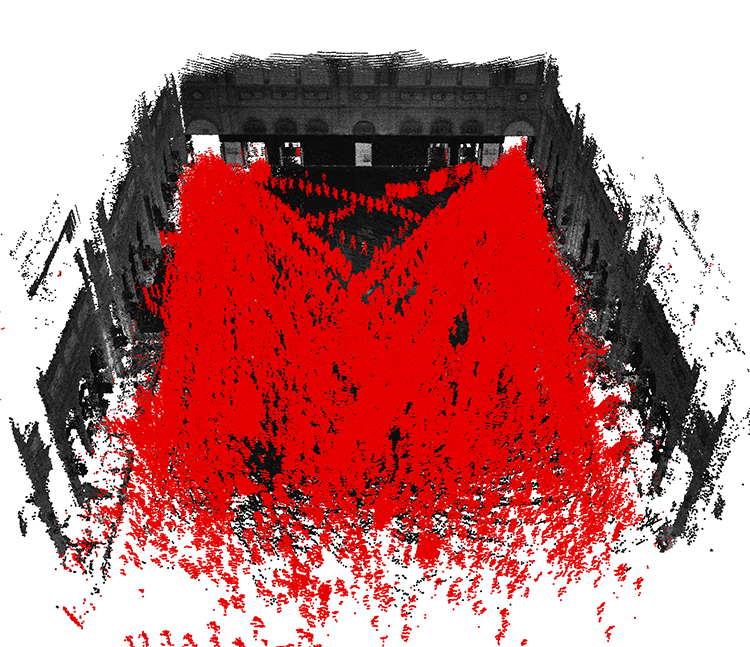}
\includegraphics[width=0.49\linewidth]{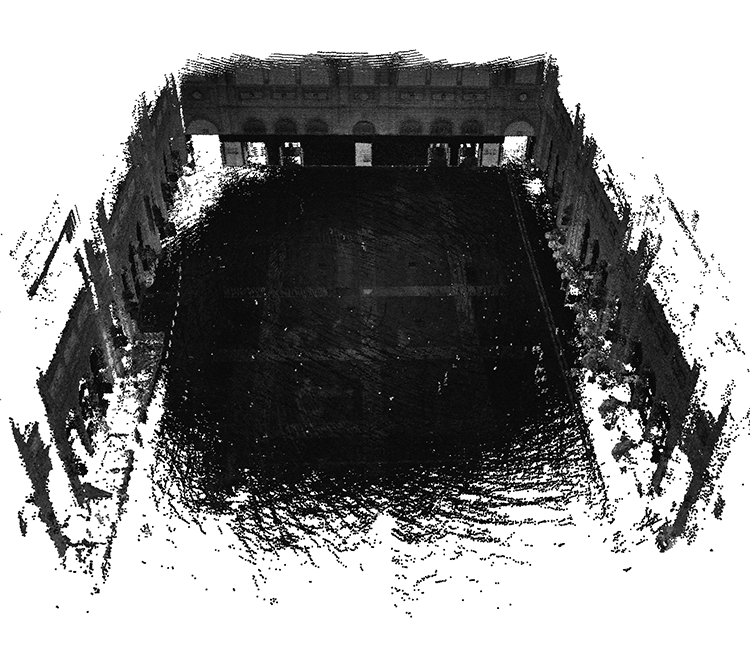}
\includegraphics[width=\linewidth]{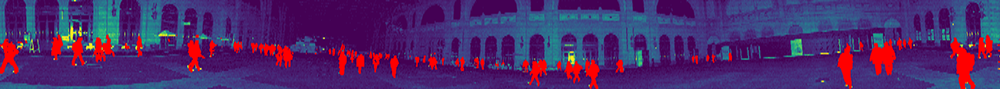}
 \vspace{-4mm}
\caption{\textit{Top:} Point cloud map built using the dynamic object aware LOAM algorithm. Static points included in the map are shown in grayscale, while points colored in red originate from dynamic objects that were discarded. \textit{Bottom:} Example output of dynamic object detections in a single 3D LiDAR scan of the same scene. Dynamic points are colored red in the intensity image resulting from the point cloud. Dynamic object detection is performed using 3D-MiniNet~\cite{3dmininet} trained on our automatically labeled dataset.
}
\label{fig:title_image}
\vspace{-5mm}
\end{figure}

Only a handful of LiDAR-based SLAM systems deal with dynamic objects explicitly.
The most common approaches~\cite{suma++,highway,lightweightsemantic} assume that dynamic objects in the scene are restricted to a distinct set of classes (e.g. cars, pedestrians, bicyclists), that can be detected using recent advances in deep learned semantic segmentation~\cite{3dmininet,rangenet++,squeezesegv2}.
These supervised learning methods require expensive and difficult to obtain training data, and therefore the capability to detect new dynamic objects remains constrained.
Other not data driven approaches use heuristics such as ignoring all objects below a certain size because they could potentially move~\cite{IMLS-SLAM}.
However, such assumptions can lead to the removal of features that are valuable for matching or 3D reconstruction.

To the best of our knowledge, we present the first unsupervised approach to a generic dynamic object aware LiDAR SLAM, using deep learning.
Even though we also use a neural network to detect dynamic objects and train it in a supervised manner, the labels and training data are generated completely automatically, thus making the overall system unsupervised.

Currently available datasets with annotated point clouds are all manually labeled.
The most extensive ones include Waymo Open~\cite{waymo}, nuScenes~\cite{nuscenes} and SemanticKITTI~\cite{semantickitti}.
These datasets were all recorded for autonomous driving applications and therefore methods trained using this data will under-perform in drastically different environments, such as indoors.
To alleviate the necessity for time-consuming manual labeling of data we present a novel dynamic object detection approach to automatically annotate point clouds.
Our approach extends the idea that dynamic objects can be detected by observing spatio-temporal changes in occupancy grids~\cite{wellhausen} with a two stage clustering and a ratio based validation check to filter outliers.
Occupancy grid based detection requires an accurate pose estimate for each point cloud observation, but at the same time LiDAR odometry can be inaccurate in the highly dynamic environments where this approach would be used.
Even though this problem renders an occupancy based approach not suitable for online filtering of dynamic objects during LiDAR SLAM, it allows us to label point clouds in an unsupervised manner.
In contrast to the aforementioned semantic datasets, our labeling approach does not make any assumption on the type of dynamic objects in the scene. 

We automatically label over 12000 LiDAR pointclouds in an environment with large amounts of pedestrians and use the resulting dataset to train 3D-MiniNet~\cite{3dmininet}.
We leverage the trained network to predict dynamic objects in point clouds in real-time and integrate it into an existing LiDAR SLAM system~\cite{loam} to create our dynamic object aware SLAM system.
We show that by filtering points from dynamic objects online, we improve the relative translational odometry error by $39.6\%$ and generate 3D reconstructions that contain drastically less non-static points as illustrated in Figure~\ref{fig:title_image}.

The main contributions of this work are:
\begin{itemize}
\item We present a full solution to creating a dynamic object aware LiDAR SLAM system, based on a deep neural network that performs online dynamic object detection. 
\item To solve the issue of generating training data we present a novel occupancy grid based approach, that includes a two stage clustering and validation step, to automatically label arbitrary dynamic objects in LiDAR point clouds.
\item In real world experiments in highly dynamic urban environments we show a clear improvement to odometry and 3D reconstruction quality.
\end{itemize}

\section{related work}
\label{sec:related_work}
Various approaches exist to detect dynamic objects in known environments, \textit{i.e.} places for which previously built maps already exist~\cite{3dmapupdate, multisession, longterm, gaussian}.
However, since these approaches assume prior knowledge of the environment, they are not suitable for online SLAM.

Approaches to dynamic object detection in unknown environments include the one proposed by Eppenberger \textit{et al.}~\cite{eppenberger} that combines semantic detections with occupancy changes from an RGBD sensor to avoid moving obstacles.
However, this approach is not directly applicable to 3D LiDAR point clouds, that lack visual information.
Approaches such as~\cite{underwood, peopleremover} successfully build static maps purely from LiDAR point clouds that contain dynamic objects.
However, they use point clouds recorded at poses that are both temporally and spatially separated, which facilitates detection since dynamic objects move a lot in between point clouds.
Our approach works on scan sequences, where dynamic objects move only slightly (or not at all) between subsequent point clouds.
The ray-tracing approach by Yoon \textit{et al.}~\cite{maplessonline} includes a clustering step, but the detections are often incomplete or contain static areas.
Our proposed approach mitigates this, by instead including a two stage clustering designed to detect more complete objects and a validation step which removes clusters containing static areas.
Dewan et al.~\cite{sceneflow} propose using rigid scene flow to detect dynamic objects.
For slowly moving objects it is hard to distinguish scene flow from noise, while our approach is not dependent on the object velocity.

In contrast to the aforementioned approaches that assume poses as given, the approaches presented in~\cite{moosmann, motionbased} perform localization and object detection jointly.
They assume that objects can be tracked in subsequent scans~\cite{moosmann} or rely on a minimum velocity assumption~\cite{motionbased} which is often not suitable for crowds of pedestrians.

While the previous approaches are based on detecting actual motion, dynamic objects can also be detected based on their appearance by leveraging recent advances in deep learning.
In urban scenarios it can be assumed that the most common dynamic objects are typically pedestrians, bicyclists and cars.
These object classes can then be detected in point clouds using deep learned semantic segmentation methods~\cite{squeezeseg,squeezesegv2,rangenet++,lunet,3dmininet, salsanext}.
However, these approaches rely on manually labeled training data, and are limited to the subset of object types that exist in available datasets~\cite{kitti,semantickitti,nuscenes,waymo}.
Automatically generated labels are used in \cite{unsupervised_grid} to learn to detect dynamic cells, but the approach operates on 2D occupancy grids with static sensors and is thus not applicable to 3D point clouds from a moving LiDAR.
Our approach makes it possible to create annotated 3D LiDAR datasets of arbitrary moving objects automatically, which extends the field of possible applications.

Systems which incorporate dynamic object awareness into LiDAR SLAM include the work of Ruchti and Burgard~\cite{dynamicprobabilities}, which uses  a neural network to predict dynamic objects but only excludes them from mapping.
Other approaches use point-wise semantic information~\cite{lightweightsemantic, suma++, highway} to treat point matches differently depending on their class.
As our approach classifies static and dynamic objects, we are able to remove dynamic features from the whole SLAM pipeline.
\section{Methodology}
\label{sec:methodology}

\begin{figure}[t]
\includegraphics[width=\linewidth]{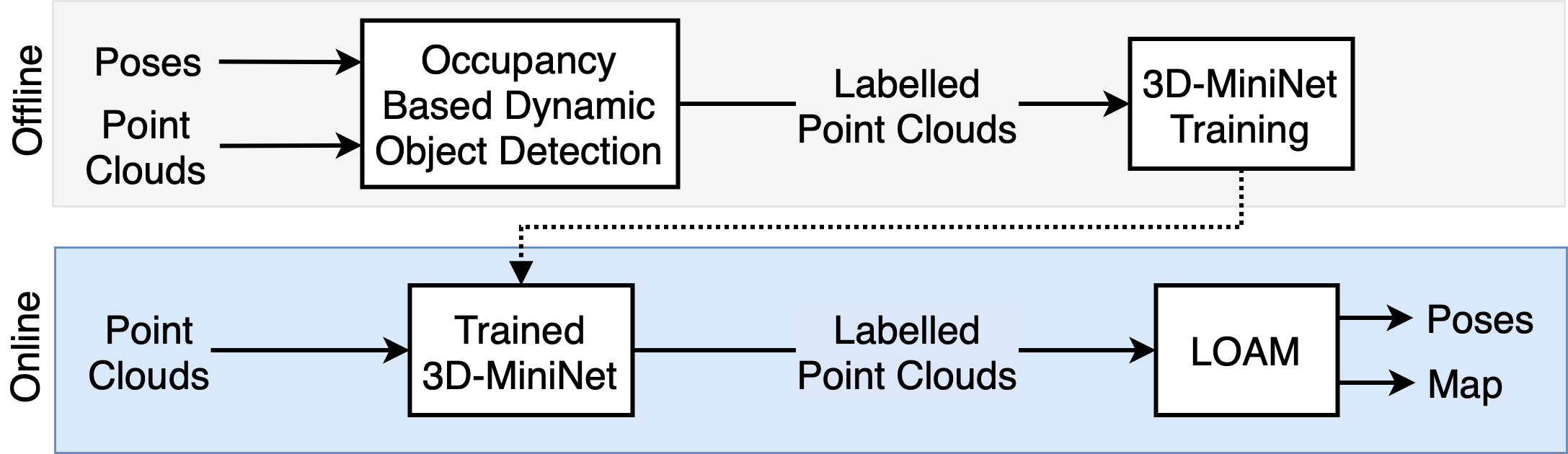}
\vspace{-6mm}
\caption{\label{fig:full_pipeline}
Full pipeline for generic dynamic object aware LiDAR SLAM.
}
\end{figure}

We present a novel occupancy grid based pipeline to detect arbitrary dynamic objects in 3D LiDAR point clouds.
This pipeline is used to generate an automatically labeled datasets of dynamic objects in an offline stage.
We then train a 3D-MiniNet network on the resulting dataset to detect dynamic objects online.
Using the deep learned online dynamic object detection we enable a dynamic object aware LiDAR SLAM pipeline, that by filtering out dynamic features achieves a more precise odometry and better 3D reconstructions of the environment.
A diagram of the full pipeline is presented in Figure~\ref{fig:full_pipeline}. 
Our contributions are focused on how we perform the occupancy grid based object detection and the proposed two stage clustering detailed in \ref{sec:eval:subsec:occupancy}.

\begin{figure}[t]
\includegraphics[width=\linewidth]{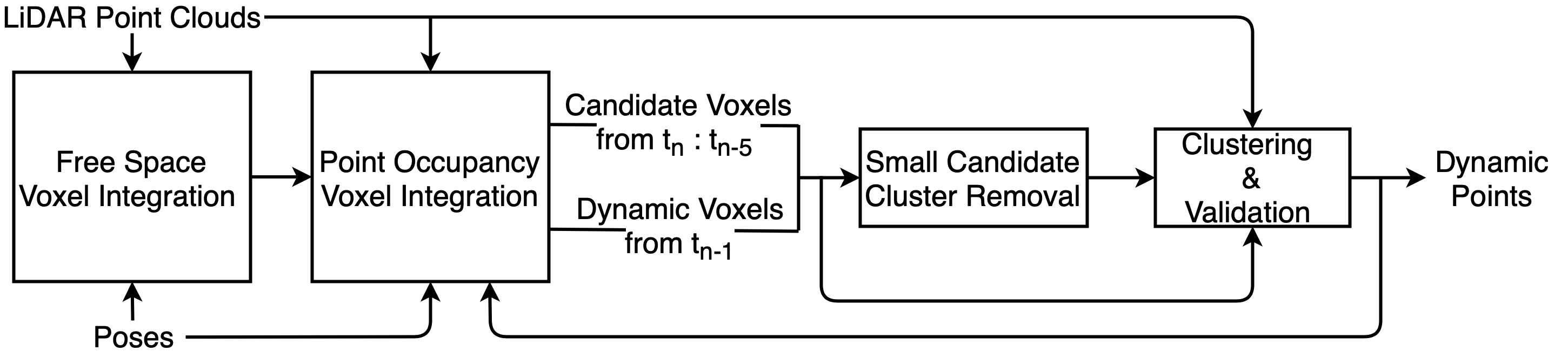}
\vspace{-5mm}
\caption{\label{fig:occupancy_pipeline}
Occupancy grid based dynamic object detection pipeline.
}
\vspace{-6mm}
\end{figure}

\subsection{Occupancy Grid Based Dynamic Object Detection}
\label{sec:methodology:subsec:occupancy_based}

\begin{figure*}[t]
\centering
\includegraphics[width=\textwidth]{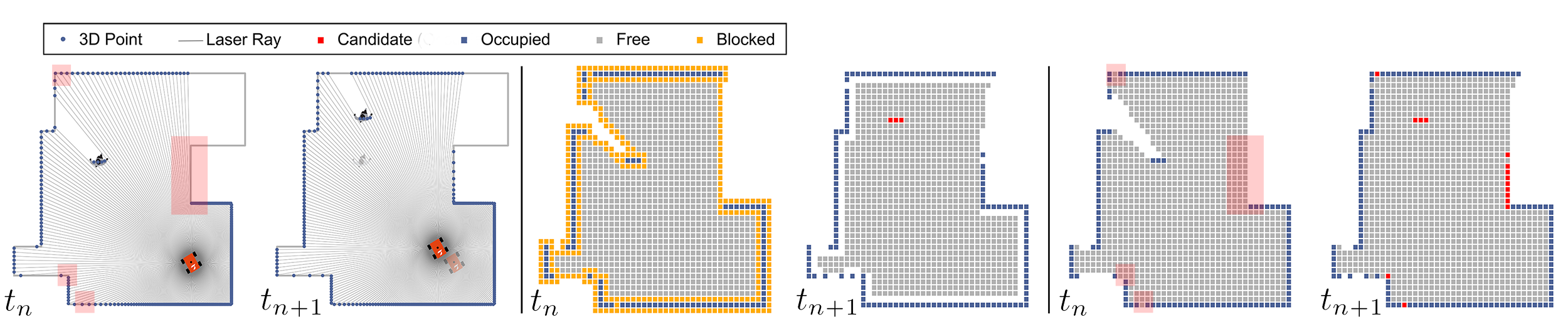}
 \vspace{-4mm}
\caption{\label{fig:occupancy_changes}
\textit{Left:}
A simplified point cloud is projected on top of a scene with walls and a pedestrian. Two different timesteps are shown. Note that the wall in the large red shaded area is occluded in the first point cloud, which results in partially occluded voxels leading to false candidate voxels for a naive ray-tracing. \textit{Middle:} Occupancy grid for our proposed approach. Voxels of the moved pedestrian are detected as candidates. Blocked voxels are not shown for the second timestep for clearness. \textit{Right:} Occupancy grid for a naive ray-tracing, where multiple false candidate voxels can be observed in red shaded areas.
}
 \vspace{-2mm}
\end{figure*}

Our proposed occupancy grid based dynamic object detection integrates successive point clouds into a global voxel grid and detects occupancy changes in this grid. In a first pass over the sequence, we acquire knowledge about all areas that have been free during the recording. We explicitly detect areas in which ray-tracing can lead to incorrect free voxels and avoid these. In the second pass, we integrate points into the previously acquired free space grid to observe occupancy changes.
If the occupancy of a voxel changes from \textit{free} to \textit{occupied}, this indicates that an object might have moved into this voxel. 
We will refer to such voxels as candidate voxels.
Purely classifying candidate voxels as dynamic points can lead to false positives and might not include all points belonging to dynamic objects.
We propose a two stage clustering and validation step to filter out noise and better detect entire objects.
A schematic of the occupancy based pipeline is given in Figure~\ref{fig:occupancy_pipeline}.
The voxel integration is implemented using Voxblox~\cite{voxblox} and a voxel size of $\unit[0.3]{m}$.

\subsubsection{Voxel Integration}
\label{sec:methodology:subsec:occupancy_based:integration}

\paragraph{Input}
The input to the voxel integration contains the points recorded during one revolution of the LiDAR.
We undistort the point cloud from the egomotion of the LiDAR by reprojecting each point using linear interpolation between the closest input poses at the pointwise timestamps.

\paragraph{Ray-Tracing}
The initial state of all voxels is \textit{unobserved}.
To detect free space, we perform ray-tracing from the LiDAR origin to the observed points.
If multiple points are observed in the same voxel, ray-tracing is only performed for the closest point to the LiDAR, to reduce the amount of redundant voxel traversals.
If a voxel that is \textit{occupied} by a currently observed point is encountered during ray-tracing, the traversal is stopped for this ray, since the area behind it is occluded.
We detect areas in which naive ray-tracing can lead to incorrect \textit{free} voxels.
Such incorrect \textit{free} voxels emerge if a ray appears along a surface that is occluded in the current scan.
In this case, the voxels containing the surface would be falsely set to \textit{free}, even though they contain objects in the real world.
This results from ray-tracing through voxels that are partially occluded due to the limited grid resolution, as outlined in Figure~\ref{fig:occupancy_changes}.
Schauer and Nüchter~\cite{peopleremover} use point normals to prevent ray-tracing through partially occluded voxels, which can be unreliable for sparse point clouds and is computationally expensive.
In contrast, we explicitly detect voxels that are partially occluded using the range image.
Partially occluded voxels arise if two points, $p_1$ and $p_2$, that are neighboring in the range image vary in range by more than $voxelsize$. In this case, the surface of the point closer to the LiDAR $p_1$, partially occludes the voxels between $p_1$ and $p_2$.
Thus, we define a discontinuity as $r_2 - r_1 > voxelsize$, where $r_n$ denotes the range of $p_n$ with $r_2 > r_1$ by definition.
In case of a discontinuity, all voxels further than $r_1$ along the ray of $p_2$ are set to \textit{blocked}. 
If a \textit{blocked} voxel is traversed, the state of it remains unchanged.
We also block all voxels that are neighboring to voxels in which a point is currently observed to add robustness to noise in the point cloud or pose.
All other traversed voxels are set to \textit{free}.

\paragraph{Free Space Pass}
If the integration would be performed by iterating over all point clouds only once in temporal order, objects that are moving into areas that are unobserved at recording time would remain undetected (e.g objects moving away in direction of the laser ray), since no assumption about the previous state could be made.
Thus, we iterate over all point clouds twice.
In the first pass, we start with an empty voxel grid and only allocate free space voxels by ray-tracing to acquire knowledge about all areas that have been free during the sequence.

\paragraph{Point Occupancy Pass}
Starting with the previously created free space voxel grid, we subsequently integrate point clouds in the second pass with a better prior knowledge on free space, thus better detecting dynamic objects. During the point occupancy pass, the same ray-tracing as in the free space pass is performed, but voxels in which points are observed are set to \textit{occupied}.
A voxel is then added to the list of candidate voxels, if its state changes from \textit{free} to \textit{occupied}.
Illustrations of examples are given in Figure~\ref{fig:occupancy_changes}.
The area occupied by a dynamic object can overlap in subsequent scans if an object moves slowly or temporarily stops moving.
In such cases, not all voxels occupied by the dynamic object would be detected as candidate voxels in two subsequent point clouds.
Thus, for the n\textit{th} LiDAR scan, at time $t_n$ we do not only consider the candidate voxels obtained based on the previous point cloud, but also include candidate voxels from timesteps $t_{n-5}$ to $t_n$.
If points have already been detected as dynamic in $t_{n-1}$, points observed in $t_n$ inside the respective voxels are considered candidate points if the voxel has been \textit{free} at least once before.
This requirement prevents the growth of false positive clusters over time.
 
\subsubsection{Two Stage Clustering}
\label{sec:methodology:subsec:occupancy_based:clustering}

\paragraph{Ground Removal}
Ground and ceiling points are removed from the candidate points by extending the approach proposed in~\cite{bogoslavskyi}.
It uses the range image to calculate an elevation angle between neighboring points in each column and assumes that each ground point segment is connected to a ground point pixel in the bottom row.
If objects are close to the LiDAR as in our datasets, this assumption is violated.
We overcome this issue by using a RANSAC plane fitting on all points with an elevation angle smaller than $30$ degree.
Restricting the plane fitting to this subset of points allows the use of a large distance threshold of $\unit[0.25]{m}$, which enables also detecting ground and ceiling points on slightly tilted or uneven surfaces.
The inlier points are then used as seed points for the ground removal clustering proposed in~\cite{bogoslavskyi}.
This assumes that the LiDAR remains approximately parallel to the ground plane.

\paragraph{Clustering}
Identified candidate points can contain false positive points from thin objects like thin poles or branches of trees, which are not reliably observed in each scan due to the sparsity of the point cloud.
In a first stage a euclidean distance based clustering using a radius of $2 * voxelsize$ is performed on the candidate points.
Only points in clusters with a diameter larger than $d = \unit[0.2]{m}$ are added to the seed points for the next stage.
The remaining seed points are used to find their respective clusters in the full point cloud.
We use the range image based clustering approach proposed in~\cite{bogoslavskyi}.
Points that are detected as ground or ceiling can be added at the boundary of a cluster, but they are not used to continue region growing.
This enables more complete object boarders close to the ground such as feet of pedestrians that are often detected as ground.

\paragraph{Candidate Cluster Validation}
If a cluster is dynamic, the majority of its points should be detected as candidate points. 
Otherwise, candidate points inside the cluster might result from noise and not from a dynamic object.
For each resulting cluster, we calculate the ratio of candidate points: $R_c = \frac{\# Candidate \ Points \ in \ Cluster}{\# Points \ in \ Cluster}$ and reject a cluster if $R_c$ is lower than $0.6$ or if the cluster contains less than $5$ points.
The parameters are hand-tuned for our sensor and scenario. 

\subsection{Dynamic Object Aware Lidar Odometry and Mapping}
\label{sec:methodology:subsec:loam}

LOAM~\cite{loam} is based on matching edge and plane features extracted by calculating the smoothness of the local surface in each point cloud. 
LOAM performs two separate scan matching steps.
Scan-to-scan matching is performed between corresponding features in subsequent scans ($\unit[10]{Hz}$), while scan-to-map matching ($\unit[1]{Hz}$) is performed between features of the current scan and features of the environment map that is gradually built.
Scan-to-map matching is more accurate, but also computationally more expensive.
The detailed algorithm is found in~\cite{loam}. 
We use a custom implementation of their work, that we refer to as \textit{standard LOAM}.
LOAM is based on the assumption, that most of the features used for matching are fixed in space, which is violated by dynamic objects.

We use the trained network described in~\ref{sec:methodology:subsec:3dmininet} to estimate dynamic object points.
Simply removing all dynamic points prior to feature detection leads to artificial edges, that are then detected as features.
Instead, we detect all features and classify them as static or dynamic. 
A feature is considered static, if all points that contribute to it were classified as static.
To maintain the assumption of a static environment, features classified as dynamic are neither used for feature matching nor added to the map.
We refer to this approach as \textit{dynamic object aware LOAM}.

\subsection{3D-MiniNet Training}
\label{sec:methodology:subsec:3dmininet}
3D-MiniNet was proposed by Alonso et al. and performs state of the art semantic segmentation for 3D LiDAR point clouds.
It consists of two main modules: The projection learning module learns a 2D representation from the $x,y,z$, intensity and range value of each point, that is then fed into the fully convolutional MiniNet~\cite{mininet} backbone that predicts a semantic label for each point.
With a runtime of $\unit[36]{fps}$ on an Nvidia RTX 2080 Ti GPU, it operates faster than the recording frequency of the sensor, which is crucial for real-time operation.
We use label smoothing~\cite{labelsmoothing} to make training more robust to false annotations that are present in our dataset.
We horizontally flip the images with a probability of $0.5$ and adapt the $x$ and $y$ values accordingly to augment the training data.
We train the network using Adam optimizer~\cite{adam} for 35 epochs with a learning rate of $0.0003$ and batch size $3$.
\section{Evaluation}
\label{sec:evaluation}
We evaluate each component of our proposed pipeline as well as the entire end-to-end process.
In a simulated environment with ground truth annotations we show that the occupancy grid based dynamic object detection is applicable to a multitude of different types of dynamic objects.
We annotated a subset of our real world dataset that we present in~\ref{sec:methodology:subsec:dataset}, to evaluate the segmentation performance of the occupancy grid based detection methods as well as of the 3D-MiniNet trained on it.
Finally, we compare the performance of a standard LOAM pipeline with our dynamic object aware LOAM.
We evaluate all 4 locations of the dataset using the same settings. At each location, 2 sequences are evaluated. Trajectory lengths vary between $\unit[100-400]{m}$ and sequences last between $\unit[100-200]{s}$. 
To evaluate 3D-MiniNet at one location, the network is trained on the data of the 3 remaining locations, e.g. we train on data from \textit{Hauptgebaeude}, \textit{Shopville} and \textit{Station} to evaluate at \textit{Niederdorf}.
Experiments were run on an Intel I7-8559U CPU and using an Nvidia Geforce RTX 2080 Ti GPU.

\subsection{Simulated Dataset}
\label{sec:evaluation:subsec:simulation}
We evaluate the performance of the occupancy grid based pipeline on a wide variety of arbitrary objects in a simulated environment of a small town.
The environment contains moving cars, planes, pedestrians, animals, cylinders, spheres and cubes at different sizes that are moving horizontally and vertically at different velocities and in different directions.
The simulated sensor is moving in a closed trajectory through the environment.
Ground truth annotations for all moving objects for each of the $1642$ point clouds are available.

\begin{figure}[t]
\centering
\begin{tabular}{@{}c@{}}
\includegraphics[width=\linewidth]{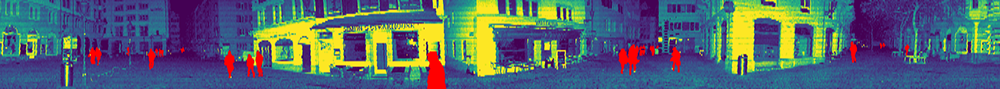}
 \end{tabular}
\begin{tabular}{@{}c@{}}
\includegraphics[width=\linewidth]{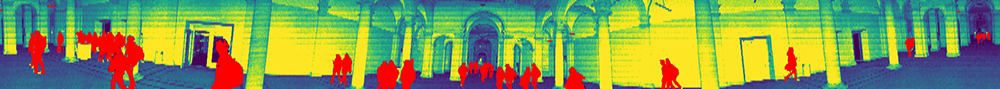}
 \end{tabular}
\begin{tabular}{@{}c@{}}
\includegraphics[width=\linewidth]{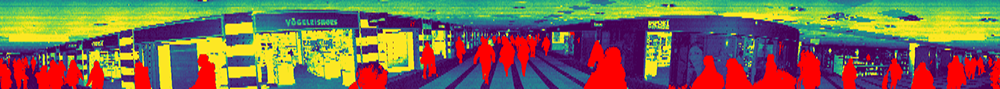}
 \end{tabular}
\begin{tabular}{@{}c@{}}
\includegraphics[width=\linewidth]{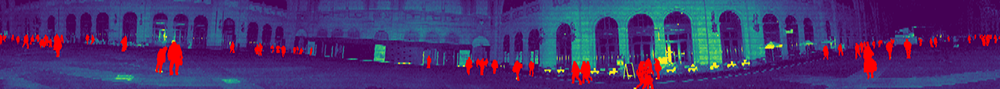}
 \end{tabular}
\vspace{-4mm}
\caption{\label{fig:dataset_examples}
Occupancy grid based detection examples shown on the intensity image generated from the point clouds. Detected dynamic objects are colored in red. \textit{Top} to \textit{Bottom}: \textit{Niederdorf, Hauptgebaeude, Shopville, Station.}
}
\vspace{-2mm}
\end{figure}

\begin{figure}[t]
\centering
\includegraphics[width=\linewidth]{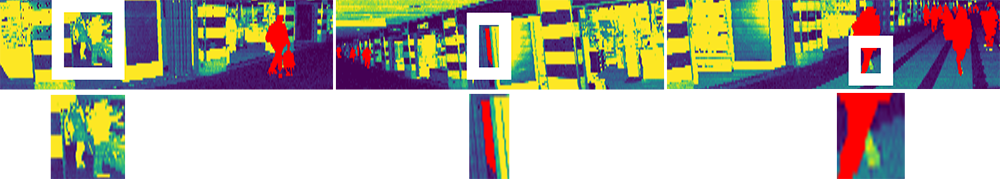}
\vspace{-6mm}
\caption{Examples of erroneous annotations in the dataset:
\textit{Left:} Missed detection from under-segmentation: Humans are close to a shelf and end up in a cluster with it.
\textit{Middle:} False positive resulting from reflective surfaces. 
\textit{Right:} The object is detected, but the lower leg is missing.
}
\label{fig:false_labels}
\vspace{-4mm}
\end{figure}

\subsection{Real World Dataset}
\label{sec:methodology:subsec:dataset}
We recorded a total of more than 12000 scans in the main hall of ETH Zurich (\textit{Hauptgebaeude}), at two different levels of the main train station in Zurich (\textit{Station}, \textit{Shopville}) and in a touristic pedestrian zone (\textit{Niederdorf}).
Examples of the collected point clouds are shown in Figure~\ref{fig:dataset_examples}.
A handheld Ouster OS1 64 LiDAR and a Alphasense Core multi-camera module sensor\footnote{\url{https://github.com/sevensense-robotics/alphasense_core_manual}} were used for recording.
Point clouds are recorded at $\unit[10]{Hz}$ with $2048$ points per revolution.
In addition a VI-SLAM pipeline is run on the Sevensense sensor data to obtain high frequency pose estimates.

We use the pipeline presented in~\ref{sec:methodology:subsec:occupancy_based} to automatically label dynamic objects in LiDAR point clouds.
For evaluation purposes we manually annotated a subset of our dataset.
Pedestrians and objects associated to them (e.g suitcases, bicycles, dogs) were annotated for 10 temporally separated point clouds for each of the 8 sequences.
Pedestrians were annotated by appearance only, thus it was not considered if they are static or moving.
\subsection{Occupancy Grid Based Dynamic Object Detection Results}
\label{sec:eval:subsec:occupancy}
The occupancy grid based dynamic object detection achieves an Intersection over Union~(IoU) of $0.92$ for moving objects averaged over all sequences of our simulated environment.
This shows, that our approach is applicable to a wide variety of objects of different dimensions and shapes and is also not restricted to a certain type of motions.
Missing detections are caused by thin or far away objects, because their resulting clusters fall below the minimum required amount of points, due to the point cloud sparsity.

On our real world dataset, an IoU of $0.88$ is achieved averaged over all locations on the annotated test set.
The error is partly due to ground truth annotation that are based on appearance. Some sequences contain pedestrians that do not move and thus are not detected by the approach.
Detailed results are provided in Table~\ref{table:iou}.
As shown in Figure~\ref{fig:dataset_examples} the vast majority of pedestrians are detected, also if they are partly occluded or close to the LiDAR.
Detection fails in some cases if a pedestrian is close to a static object, as points belonging to the pedestrian end up in a cluster with the static object and thus the cluster validation is not passed.
In rare cases small parts like a leg of a pedestrian remain undetected due to over-segmentation.
The detection performance decreases with distance from the sensor, due to the increasing point sparsity.
A limitation of LiDAR sensors is that reflective surfaces can cause invalid distance measurements, that lead to invalid voxel states and erroneous annotations.
Examples of erroneous annotations are given in Figure~\ref{fig:false_labels}.
The detection takes $\unit[1.2]{s}$ on average per point cloud. 

\begin{figure}[t]
\centering
\begin{tabular}{@{}c@{}}
\includegraphics[width=\linewidth]{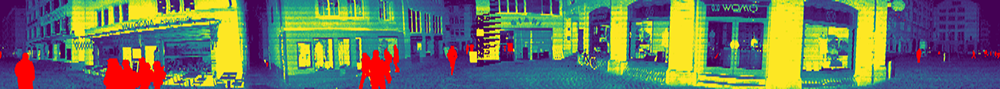}
 \end{tabular}
\begin{tabular}{@{}c@{}}
\includegraphics[width=\linewidth]{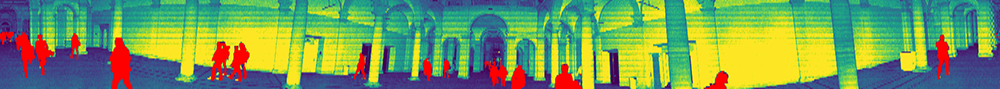}
 \end{tabular}
\begin{tabular}{@{}c@{}}
\includegraphics[width=\linewidth]{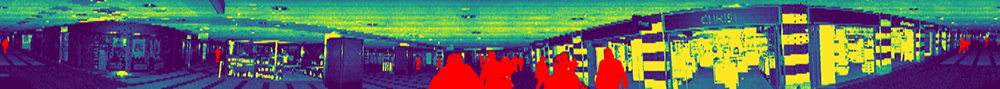}
 \end{tabular}
\begin{tabular}{@{}c@{}}
\includegraphics[width=\linewidth]{images/3dm-hb-ground-example.png}
\end{tabular}
 \vspace{-4mm}
\caption{3D-MiniNet prediction examples at the four different locations. For each location, the training was performed on the automatically labeled data of the 3 remaining locations.}
\vspace{-2.5mm}
\label{fig:3dm_examples}
\end{figure}

\begin{figure}[t]
\centering
\begin{tabular}{@{}c@{}}
\includegraphics[width=\linewidth]{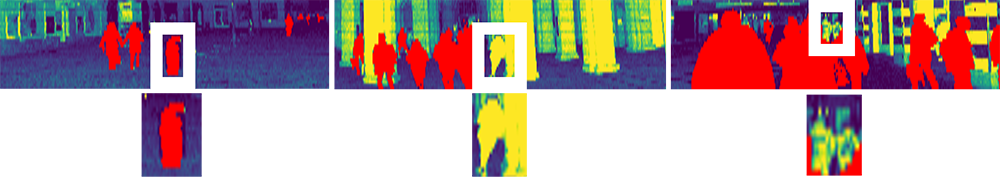}
 \end{tabular}
  \vspace{-3mm}
\caption{Erroneous 3D-MiniNet Prediction Results:
\textit{Left:} False positive detection on a trash bin.
\textit{Middle:} Missed detection on a person leaning onto a pillar with similar intensity values.
\textit{Right:} False negative prediction on two pedestrians in a cluttered group.
}
\label{fig:false_predictions}
\vspace{-5mm}
\end{figure}

\begin{table}[b] 
\begin{center}
\caption{
Pedestrian IoU on test set at different locations 
}\label{table:iou}
\begin{adjustbox}{max width=\linewidth}

\begin{tabular}{ccccc} 
\toprule
\backslashbox{\tabular{@{}l@{}}Method\endtabular}{Location} & Station & Shopville & Hauptgebaeude & Niederdorf \\
\midrule
\multicolumn{1}{l}{\textbf{\emph{Occupancy Grid}}} & 0.91 & 0.85 & 0.88 & 0.87\\
\\
\multicolumn{1}{l}{\textbf{\emph{3D-MiniNet}}} & 0.84 & 0.82 & 0.82 & 0.8\\ \bottomrule
\end{tabular}
\end{adjustbox}
\end{center}
\vspace{-1.5mm}
\end{table}

\subsection{3D-MiniNet Based Dynamic Object Detection Results}
\label{sec:evaluation:subsec:3dmininet}
 We achieve an IoU of $0.82$ for the pedestrian class, averaged over all locations on the manually obtained ground truth annotations for the real world dataset.
Results for the individual locations are provided in Table~\ref{table:iou}. Examples of the detections are given in Figure~\ref{fig:3dm_examples}. 
%
The performance decreases with increasing point distance, as is to be expected, as more missing annotations are present at higher distance.
\begin{figure}[t]   
\captionsetup[subfigure]{labelformat=empty}

\includegraphics[width=\linewidth]{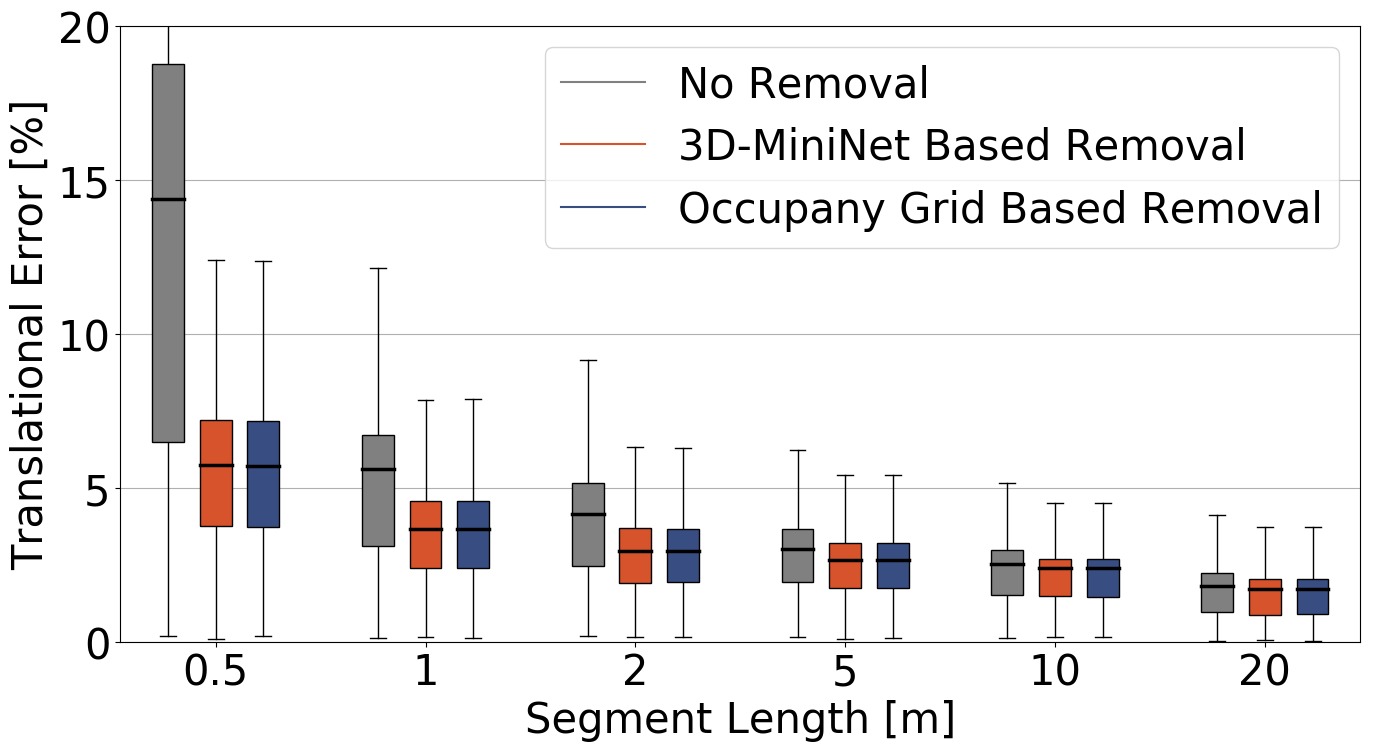}
 \vspace{-6mm}  
\caption{
Relative trajectory errors for different segment lengths. Mean values are indicated by black bars.
}
\label{fig:odom_results}
 \vspace{-2mm}  
\end{figure}
\begin{figure}[t]
\includegraphics[width=\linewidth]{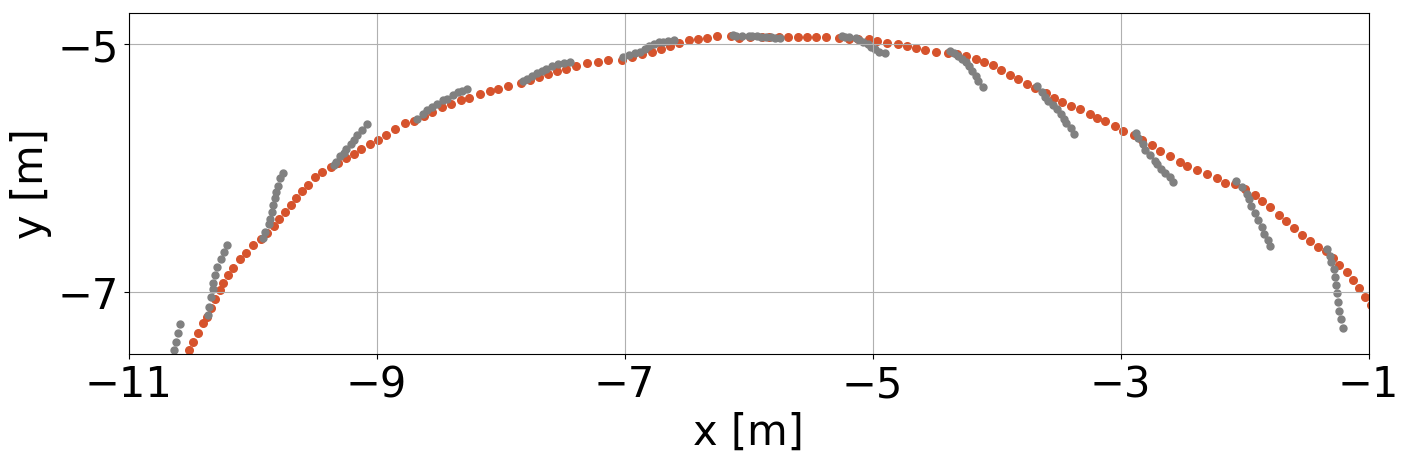}
\vspace{-7mm}
\caption{\label{fig:trajectories-compared} Trajectory segment of sequence \textit{Station-2}: The trajectory from our standard LOAM implementation(grey) contains several gaps, in contrast to the trajectory obtained from the dynamic object aware LOAM (orange).
}
\vspace{-3mm}
\end{figure}
\subsection{Odometry Results}
We compare the trajectories resulting from dynamic object aware LOAM to standard LOAM.
We also show the results that would be achieved by using dynamic object aware LOAM with the pointwise labels from the occupancy based dynamic object detection as a reference. 
This pipeline can not be used online, as the occupancy based detection relies on previously known pose estimates and free space knowledge acquired by future observation.
We use the globally bundle adjusted poses of the VI-SLAM pipeline as a ground truth reference to evaluate the odometry.

We calculate the relative trajectory error as in~\cite{kitti}, for overlapping segments of $0.5$, $1$, $2$, $20$ and $\unit[20]{m}$, which we deem representative for applications in the given environments.
We average the respective errors over all sequences.

Estimating odometry using the dynamic object aware LOAM improves the relative translational error on all segments lengths compared to standard LOAM. Averaged over all segments, $39.6\%$ less translational drift is achieved. The results are presented in more detail in Figure~\ref{fig:odom_results}. 
The translational error decreases especially for shorter segments. This mainly results from drift in the standard LOAM approach during scan-to-scan steps. This drift is mainly compensated during scan-to-map steps and thus has a lower effect on longer segments.
Looking qualitatively at the trajectories we observe that the bias in the drift correlates with the general direction of movement of nearby people in the dataset.
This is also reflected in the trajectories estimated by both methods as shown in an example trajectory segment in Figure~\ref{fig:trajectories-compared}. It can be seen that the standard LOAM has very clear low frequency discontinuities in the odometry of $\unit[0.14]{m}$ on average, whenever the more accurate scan matching is performed.
This non smooth odometry poses significant disadvantages when used for navigation or obstacle avoidance.
In contrast, the trajectory of dynamic object aware LOAM is much smoother and reduces the low frequency jumps to  $\unit[0.04]{m}$ on average, making it a much better candidate for use on a robotic platform.
The relative rotational error remained approximately equal across approaches in our experiments.

\subsection{Mapping}
\begin{figure} [t]
\vspace{-2mm}
\captionsetup[subfigure]{labelformat=empty}

       \subfloat[\label{1b}]{%
        \includegraphics[width=0.48\linewidth]{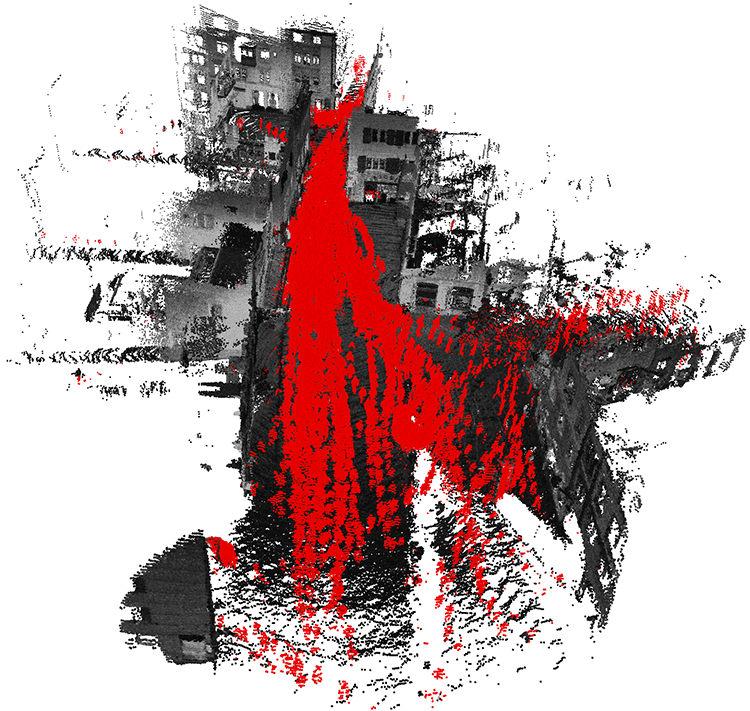}}
    \hfill
  \subfloat[]{%
        \includegraphics[width=0.48\linewidth]{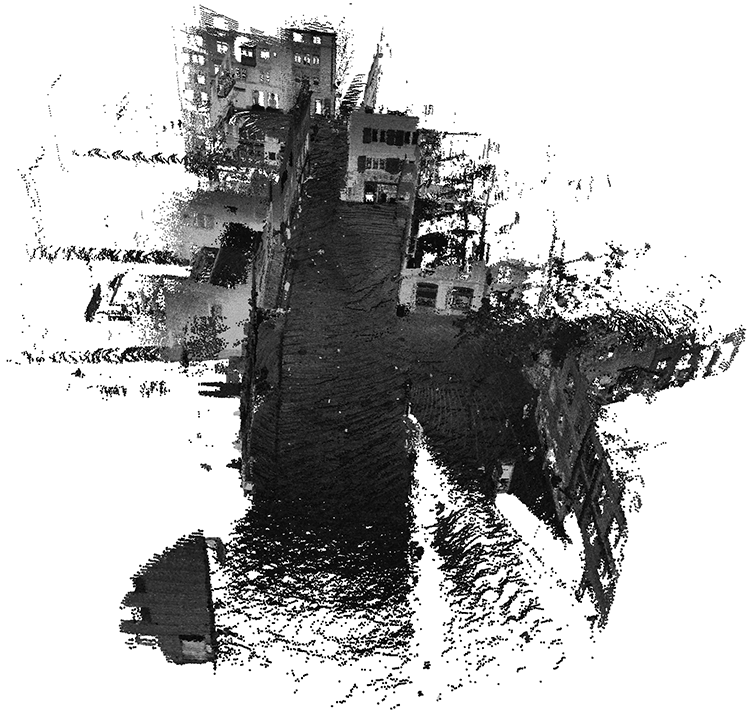}}
   
  \label{fig1} 
   \vspace{-2mm}
\caption{Resulting map for sequence \textit{Niederdorf-1}: \textit{Left:} Points classified as \textit{dynamic} are shown in red, \textit{static} points are shown in grayscale. \textit{Right:} The cleaned map without points classified as \textit{dynamic} in grayscale.}
\label{fig:maps}
\vspace{-3mm}
\end{figure}

Maps presented in Figures~\ref{fig:maps} and~\ref{fig:title_image} were built by aggregating point clouds using poses obtained from the scan-to-map matching steps and filtering out points that are further than $\unit[30]{m}$ from the LiDAR.
The maps were subsampled using a $\unit[0.1]{m}$ voxel grid.
Removing dynamic objects makes the static structure in the scene far more clearly observable.

\section{conclusions}
\label{sec:conclusions}
We presented a complete solution for creating a dynamic object aware LiDAR SLAM pipeline, that is based on a deep learned dynamic object filtering step.
To this end we proposed a novel occupancy grid based approach to automatically label arbitrary dynamic objects in point clouds offline, to efficiently create training data for any dynamic environment.
We leveraged our method to automatically annotate a large amount of pedestrians and other dynamic objects at four distinct, highly dynamic, urban locations in more than 12000 real world LiDAR point clouds. 
The dataset is then used to train a 3D-MiniNet neural network to segment out dynamic objects in real-time and enhance the performance of LOAM by removing these objects from the point clouds before the matching and mapping process.
This improves odometry by reducing drift on average by $39.6\%$ and also significantly smoothing out the trajectory estimate.
In addition we are able to create much better and accurate 3D scene reconstructions.
Even though we showcase our proposed pipeline using 3D-MiniNet and LOAM as well as a dataset with a significant amount of pedestrians, our proposed methods are generic and equally applicable to a large variety of dynamic objects, segmentation methods and LiDAR SLAM algorithms.

\clearpage

\addtolength{\textheight}{0pt}   








\bibliographystyle{IEEEtran}
\bibliography{mendeley_2}

\end{document}